\pdfoutput=1
\documentclass[11pt]{article}

\usepackage{ACL2023}

\usepackage{times}
\usepackage{latexsym}

\usepackage[T1]{fontenc}
\usepackage[utf8]{inputenc}

\usepackage{microtype}

\usepackage{inconsolata}

\usepackage[ruled,linesnumbered,vlined]{algorithm2e}  
\usepackage{amsfonts,amsmath}
\usepackage{balance}
\usepackage{booktabs}
\usepackage{graphicx}
\usepackage{multirow}
\usepackage{paralist}
\usepackage{subfigure}

\title{Serial Contrastive Knowledge Distillation for Continual Few-shot\\ Relation Extraction}

\author{
    Xinyi Wang$^\dagger$ \quad 
    Zitao Wang$^\dagger$ \quad 
    Wei Hu$^{\dagger,\,\ddagger,\,}$\thanks{\,\, Corresponding author} \\
    $^\dagger$ State Key Laboratory for Novel Software Technology, Nanjing University, China \\
    $^\ddagger$ National Institute of Healthcare Data Science, Nanjing University, China \\
    \texttt{\{xywang,ztwang\}.nju@gmail.com, whu@nju.edu.cn} 
}

\begin{document}
\maketitle

\begin{abstract}
Continual few-shot relation extraction (RE) aims to continuously train a model for new relations with few labeled training data, of which the major challenges are the catastrophic forgetting of old relations and the overfitting caused by data sparsity. 
In this paper, we propose a new model, namely SCKD, to accomplish the continual few-shot RE task. 
Specifically, we design serial knowledge distillation to preserve the prior knowledge from previous models and conduct contrastive learning with pseudo samples to keep the representations of samples in different relations sufficiently distinguishable. 
Our experiments on two benchmark datasets validate the effectiveness of SCKD for continual few-shot RE and its superiority in knowledge transfer and memory utilization over state-of-the-art models.
\end{abstract}

%====================%
\section{Introduction}
\label{sect:intro}

Relation extraction (RE) aims to recognize the semantic relations between entities in texts, which is widely applied in many downstream tasks such as language understanding and knowledge graph construction. 
Conventional studies \cite{zeng2014relation,heist2017language,zhang2018graph} mainly assume a fixed pre-defined relation set and train on a fixed dataset. 
However, they cannot work well with the new relations that continue emerging in some real-world scenarios of RE. 
Continual RE \cite{wang2019sentence,han2020continual,wu2021curriculum} was proposed as a new paradigm to solve this situation, which applies the idea of continual learning \cite{parisi2019continual} to the field of RE. 

Compared with conventional RE, continual RE is more challenging. 
It requires the model to learn emerging relations while maintaining a stable and accurate classification of old relations, i.e., the so-called \emph{catastrophic forgetting} problem \cite{thrun1995lifelong,french1999catastrophic}, which refers to the severe loss of prior knowledge during the model is learning new tasks. 
Recent continual learning works leverage the regularization-based models, the architecture-based models, and the memory-based models to alleviate catastrophic forgetting. 
Several studies \cite{wang2019sentence,sun2020lamol} have shown that the memory-based models are more promising for NLP tasks, and a number of memory-based continual RE models \cite{cui2021refining,zhao2022consistent,hu2022improving,zhang2022prompt} have made significant progress.

\begin{figure}[!tb]
\centering
\includegraphics[width=.98\columnwidth]{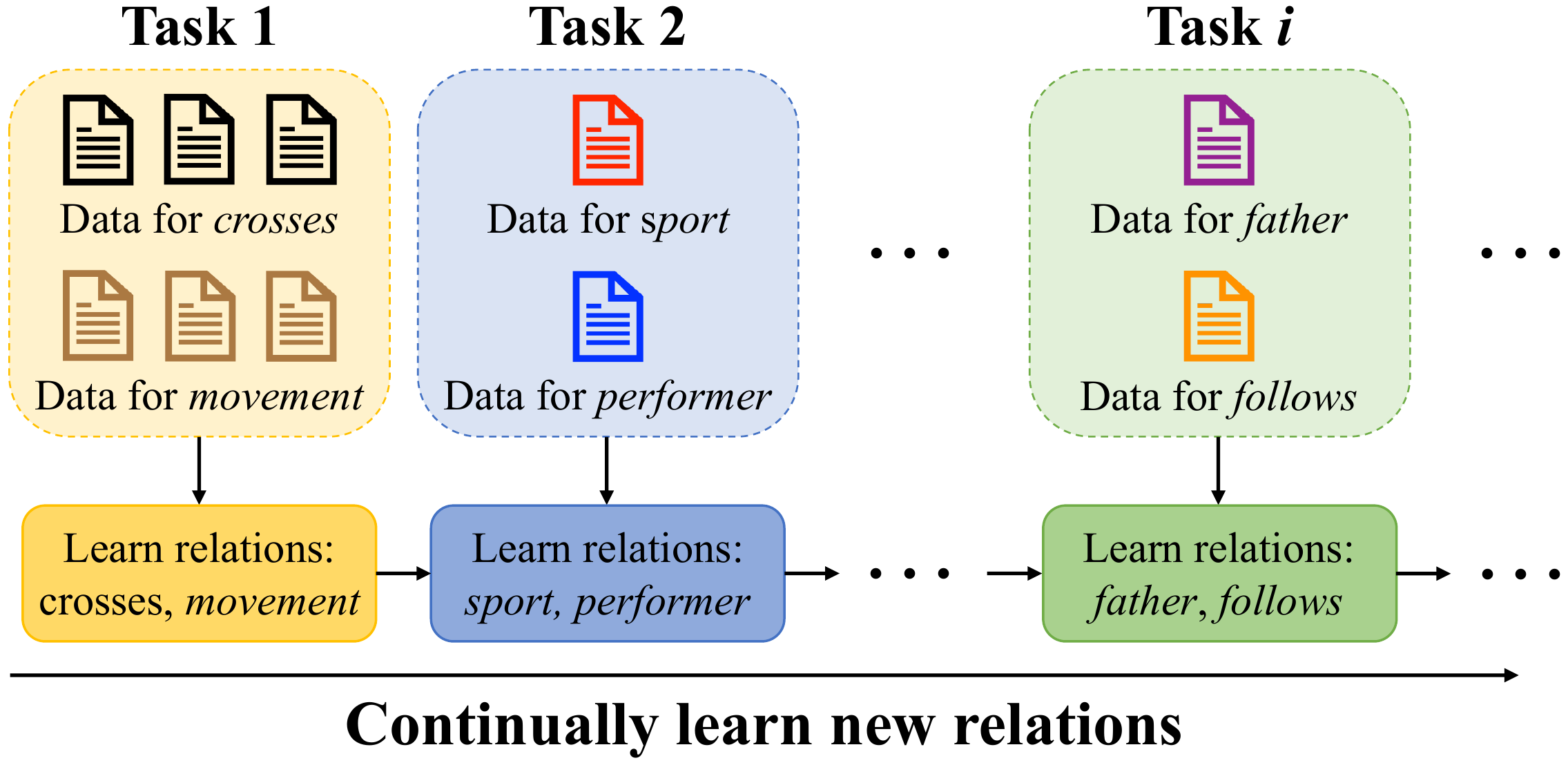}
\caption{Continual few-shot RE paradigm}
\label{fig:example}
\end{figure}

In real life, the shortage of labeled training data for relations is an unavoidable problem, especially severe in emerging relations. 
Therefore, the \emph{continual few-shot RE} paradigm \cite{qin2022continual} was proposed to simulate real human learning scenarios, where new knowledge can be acquired from a small number of new samples. 
As illustrated in Figure~\ref{fig:example}, the continual few-shot RE paradigm expects the model to continuously learn new relations through abundant training data only for the first task, but through sparse training data for all subsequent tasks.
Thus, the model needs to identify the growing relations well with few labeled data for them while retaining the knowledge on old relations without re-training from scratch.
As relations grow, the confusion about relation representations leads to catastrophic forgetting. 
In continual few-shot RE, catastrophic forgetting becomes more severe since the few samples of new relations may not be representative for these relations. 
The possibility of confusion between relation representations greatly increases.
Since the emerging relations are few-shot, the problem of \emph{overfitting} becomes another key challenge in the continual few-shot RE task. 
The overfitting for samples in few-shot tasks aggravates the model's forgetting of prior knowledge as well. 
Existing few-shot learning works \cite{fan2019large,gao2019hybrid,obamuyide2019model,geng2020mick} are worthy of reference by continual few-shot RE models to ensure good generalization.

Inspired by knowledge distillation \cite{hinton2015distilling} to transfer knowledge well and contrastive learning \cite{wu2018unsupervised} to constrain representations explicitly, we propose SCKD, a model built with \emph{serial contrastive knowledge distillation} for continual few-shot RE. 
Through it, we tackle the aforementioned two key challenges. 
First, how to alleviate the problem of catastrophic forgetting? 
SCKD follows the memory-based methods for continual learning and preserves a few typical samples from previous tasks.
Furthermore, we present serial knowledge distillation to preserve the prior knowledge from previous models and conduct contrastive learning to keep the representations of samples in different relations sufficiently distinguishable.
Second, how to mitigate the negative impact of overfitting caused by sparse samples?
We leverage bidirectional data augmentation between memory and current tasks to obtain more samples for few-shot relations. 
The pseudo samples generated in serial contrastive knowledge distillation can help prevent overfitting as well. 

In summary, our main contributions are twofold:
\begin{itemize}
\item We propose SCKD, a novel model built with serial contrastive knowledge distillation for resolving the continual few-shot RE task.
With the proposed serial knowledge distillation and contrastive learning with pseudo samples, our SCKD can take full advantage of memory and effectively alleviate the problems of catastrophic forgetting and overfitting under considerably few memorized samples.

\item We perform extensive experiments on two benchmark datasets FewRel \cite{han2018fewrel} and TACRED \cite{zhang2017position}. 
The results demonstrate the superiority of SCKD over the state-of-the-art continual (few-shot) RE models.
Furthermore, the proposed data augmentation, serial knowledge distillation, and contrastive learning all contribute to performance improvement.
\end{itemize}

%====================%
\section{Related Work}
\label{sect:work}

In this section, we review related work on continual RE and few-shot RE.

\paragraph{Continual RE.}
The goal of continual learning is to accomplish new tasks sequentially without catastrophically forgetting the acquired knowledge from previous tasks.  
For continual RE, RP-CRE \cite{cui2021refining} refines sample embeddings for prediction with the generated relation prototypes from memory. 
However, its relation prototype calculation is sensitive to typical samples.
CRL \cite{zhao2022consistent} introduces supervised contrastive learning and knowledge distillation to generate sample representations when replaying memory. 
It narrows the representations of samples belonging to the same relation through supervised contrastive learning but fails to keep the representations of samples in different relations far away to avoid confusion. 
Besides, knowledge distillation between prototypes calculated by averaging sample representations may lose some features of specific samples.
CRECL \cite{hu2022improving} contrasts a given sample with all the candidate relation prototypes stored in memory by a contrastive network. 
It faces the same problem as RP-CRE on typical samples for computing relation prototypes. 
Conducting contrastive learning only with relation prototypes may not guarantee the differences between sample representations belonging to different relations.
KIP-Framework \cite{zhang2022prompt} generates knowledge-infused relation prototypes to leverage the relational knowledge from pre-trained language models with prompt tuning. 
Compared with other models, KIP-Framework needs extra knowledge such as relation descriptions, and its overall procedure is more time-consuming.
All these works rely on plenty of training data for learning new relations and large memory for retaining prior knowledge. 
In contrast, our model only needs a few training samples to learn new relations well through bidirectional data augmentation and the generated pseudo samples from relation prototypes.
Furthermore, our model can avoid catastrophic forgetting under limited memory through serial knowledge distillation and contrastive learning.

As far as we know, ERDA \cite{qin2022continual} is the only work addressing continual few-shot RE. 
It imposes relational constraints in the embedding space and generates new training data from unlabeled text. 
However, our model does not need to import extra data like ERDA. 
Instead, it generates pseudo samples from relation prototypes and augments training data by modifying original samples to alleviate the overfitting problem.

\paragraph{Few-shot RE.} 
Few-shot learning aims to leverage only a few novel samples to adapt the model for solving tasks. 
For few-shot RE, its goal is to enable the model to quickly learn the characteristics of relations with very few samples, so as to accurately classify these relations.
At present, there are two main lines of work:
(1) The metric learning methods \cite{fan2019large,gao2019hybrid} use various metric functions (e.g., the Euclidean or Cosine distance) learned from prior knowledge to map the input into a subspace so that they can distinguish similar and dissimilar sample pairs easily to assign the relation labels.
(2) The meta-learning methods \cite{obamuyide2019model,geng2020mick} learn general relation classification experience from the meta-training stage and leverage the experience to quickly converge on specific relation extraction during the meta-testing stage.
In this paper, our problem setting is different from the above few-shot RE works, as we expect the model to continuously learn new few-shot relations instead of conducting the few-shot relation learning just once. 
Furthermore, these few-shot RE works do not have the capacity for continual learning.

%====================%
\section{Methodology}
\label{sect:method}

% In this section, we first introduce the studied task. 
% Then, we present our model in detail. 

%--------------------%
\subsection{Task Definition}
\label{subsec:Task}

The objective of RE is to identify the relations between entity mentions in sentences. 
Continual RE aims to accomplish a sequence of $J$ RE tasks $\{T_1,T_2,\dots,T_J\}$, where each task $T_j$ has its own dataset $D_j$ and relation set $R_j$. 
The relation sets of different tasks are disjoint. 
Once finishing $T_j$, $D_j$ is no longer available for future learning, and the model is assessed on all previous tasks $\{T_1,$ $\dots,T_j\}$ for identifying $\tilde{R}_j=\bigcup_{i=1}^j R_i$. 
Also, the trained model serves as the base model for the subsequent task $T_{j+1}$.

In real-world scenarios, labeled training data for new tasks are often limited. 
Therefore, we define the continual few-shot RE task in this paper, where only the first task $T_1$ has abundant data for model training and the subsequent tasks are all few-shot. 
Let $N$ be the relation number of each few-shot task and $K$ be the sample number of each relation, the task can be called \emph{$N$-way-$K$-shot}. 
A continual few-shot RE model is expected to perform well on all historical few-shot and non-few-shot tasks.

%--------------------%
\subsection{Our Framework}
\label{subsec:Framework}

Algorithm \ref{alg:Framework} shows the end-to-end training for task $T_j$, with the model $\Phi_{j-1}$ previously trained.
Following the memory-based methods for continual learning \cite{lopezpaz2017advances,chaudhry2019efficient}, we use a memory $\tilde{M}_{j-1}$ to preserve a few samples in all previous tasks $\{T_1,\dots,T_{j-1}\}$.
\begin{enumerate}
    \item \textbf{Initialization} (Line 1). 
    The current model $\Phi_j$ inherits the parameters of $\Phi_{j-1}$, except for $\Phi_1$ randomly initialized. 
    We adapt $\Phi_j$ on $D_j$ to learn the knowledge of new relations in $T_k$.
    
    \item \textbf{Prototype generation} (Lines 2--6). 
    Inspired by \cite{han2020continual,cui2021refining}, we apply the k-means algorithm to select $L$ typical samples from $D_j$ for every relation $r\in R_j$, which constitute a memory $M_r$. 
    The memory for the current task is $M_j=\bigcup_{r\in R_j}M_r$, and the overall memory for all observed relations until now is $\tilde{M}_j=\tilde{M}_{j-1}\cup M_j$. 
    Then, we generate a prototype $\mathbf{p}_r$ for each $r\in\tilde{R}_j$.
    
    \item \textbf{Data augmentation } (Line 7). 
    To cope with the scarcity of samples, we conduct bidirectional data augmentation between $D_j$ and $\tilde{M}_j$. By measuring the similarity between entities in samples, we generate an augmented dataset $D_j^*$ and an augmented memory $\tilde{M}_j^*$ by mutual replacement between similar entities. 
    
    \item \textbf{Serial Contrastive Knowledge Distillation} (Lines 8--10). 
    We construct a set of pseudo samples based on the prototype set. Then, we carry out serial contrastive knowledge distillation with the pseudo samples on $D_j^*$ and on $\tilde{M}_j^*$, respectively, making the sample representations in different relations distinguishable and preserve the prior knowledge for identifying the relations in previous tasks well.
\end{enumerate}
We detail the procedure in the subsections below.

\begin{algorithm}[!t]
\caption{Training procedure for $T_j$}  
\label{alg:Framework}  
    \KwIn{$\Phi_{j-1},\tilde{R}_{j-1},\tilde{M}_{j-1},D_j,R_j$}
    \KwOut{$\Phi_j,\tilde{M_j}$}
    initialize $\Phi_j$ from $\Phi_{j-1}$, and adapt it on $D_j$\;
    $\tilde{M}_j \leftarrow \tilde{M}_{j-1}$\;
    \ForEach{$r \in R_j$}{
        pick $L$ samples in $D_j$, and add into $\tilde{M}_j$\;
    }
    $\tilde{R}_j \leftarrow \tilde{R}_{j-1} \cup R_j$\;
    generate prototype set $\tilde{P}_j$ based on $\tilde{M}_j$\;
    generate augmented dataset $D_j^*$ and memory $\tilde{M}_j^*$ by mutual replacement\;
    generate pseudo sample set $\tilde{S}_j$ based on $\tilde{P}_j$\;
    update $\Phi_j$ by serial contrast. knowl. distill. on $D_j^*,\tilde{S}_j$\tcp*{\small re-train current task}
    update $\Phi_j$ by serial contrast. knowl. distill. on $\tilde{M}_j^*,\tilde{S}_j$\tcp*{\small memory replay}
\end{algorithm} 

%--------------------%
\subsection{Initialization for New Task}
\label{subsec:Initialization}

To adapt the model for the new task $T_j$, we perform a simple multi-classification task on dataset $D_j$.

Specifically, for a sample $x$ in $T_j$, we use special tokens $[E_1]$ and $[E_2]$ to denote the start positions of two entities in $x$, respectively. 
Then, we obtain the representations of special tokens using the BERT encoder \cite{devlin2019bert}.
Next, the feature of sample $x$, denoted by $\mathbf{f}_x$, is defined as the concatenation of token representations of $[E_1]$ and $[E_2]$. 
We obtain the hidden representation $\mathbf{h}_x$ of $x$ as 
\begin{align}
\mathbf{h}_x = \mathrm{LN}(\mathbf{W}\,\mathrm{Dropout}(\mathbf{f}_x)+\mathbf{b}),
\end{align}
where $\mathbf{W}\in\mathbb{R}^{d\times 2h}$ and $\mathbf{b}\in\mathbb{R}^d$ are two trainable parameters. 
$d$ is the dimension of hidden layers. 
$h$ is the dimension of BERT hidden representations.
$\mathrm{LN}(\cdot)$ is the layer normalization operation.

Finally, based on $\mathbf{h}_x$, we use the linear softmax classifier to predict the relation label. 
The classification loss, $\mathcal{L}_\text{csf}$, is defined as
\begin{align}
\mathcal{L}_\text{csf} = -\frac{1}{|D_j|} \sum_{x \in D_j}\sum_{r=1}^{|R_j|} y_{x,r} \cdot\log P_{x,r},
\end{align}
where $y_{x,r}\in\{0,1\}$ indicates whether $x$'s true label is $r$. 
$P_{x,r}$ denotes the $r$-th entry in $x$'s probability distribution calculated by the classifier.

%--------------------%
\subsection{Prototype Generation}
\label{subsec:Prototype}

After the initial adaption above, we pick $L$ typical samples for each relation $r\in R_j$ to form memory $M_r$. 
We leverage the k-means algorithm upon the hidden representations of $r$'s samples, where the number of clusters equals the number of samples that need to be stored for representing $r$. 
Then, in each cluster, the sample closest to the centroid is chosen as one typical sample.

To obtain the prototype $\mathbf{p}_r$ for $r$, we average the hidden representations of $L$ typical samples in $M_r$:
\begin{align}
\mathbf{p}_r = \frac{1}{L}\sum_{x \in M_r} \mathbf{h}_x.
\label{equ:proto}
\end{align}

The prototype set $\tilde{P}_j$ stores the prototypes of all relations in $\tilde{R}_j$, i.e., $\tilde{P}_j=\cup_{r\in \tilde{R}_j}\{\mathbf{p}_r\}$.

%--------------------%
\subsection{Bidirectional Data Augmentation}
\label{subsec:Augmentation}

For a sample $x$ in $D_j$ or $\tilde{M}_j$, the token representations of $[E_{1}]$ and $[E_{2}]$ generated by BERT are used as the representations of corresponding entities. 
We obtain the entity representations from all samples and calculate the cosine similarity between the representations of any two different entities. 
Once the similarity exceeds a threshold $\tau$, we replace each of the two entities in the original sample with the other entity. 
Our intuition is that one certain entity in a sentence is replaced by its close entity with everything else unchanged, the relation represented by the sentence is unlikely to change much. 
For example, ``The route crosses the Minnesota River at the Cedar Avenue Bridge.'' and ``The route crosses the River MNR at the Cedar Avenue Bridge.'' have the same relation ``\textit{crosses}''.
We assign the same relation label to the new samples as their original samples and store them together as the augmented dataset $D_j^*$ and the augmented memory $\tilde{M}_j^*$.

%--------------------%
\subsection{Serial Contrastive Knowledge Distillation}
\label{subsec:Distillation}

\begin{figure}[!tb]
\centering
\includegraphics[width=\columnwidth]{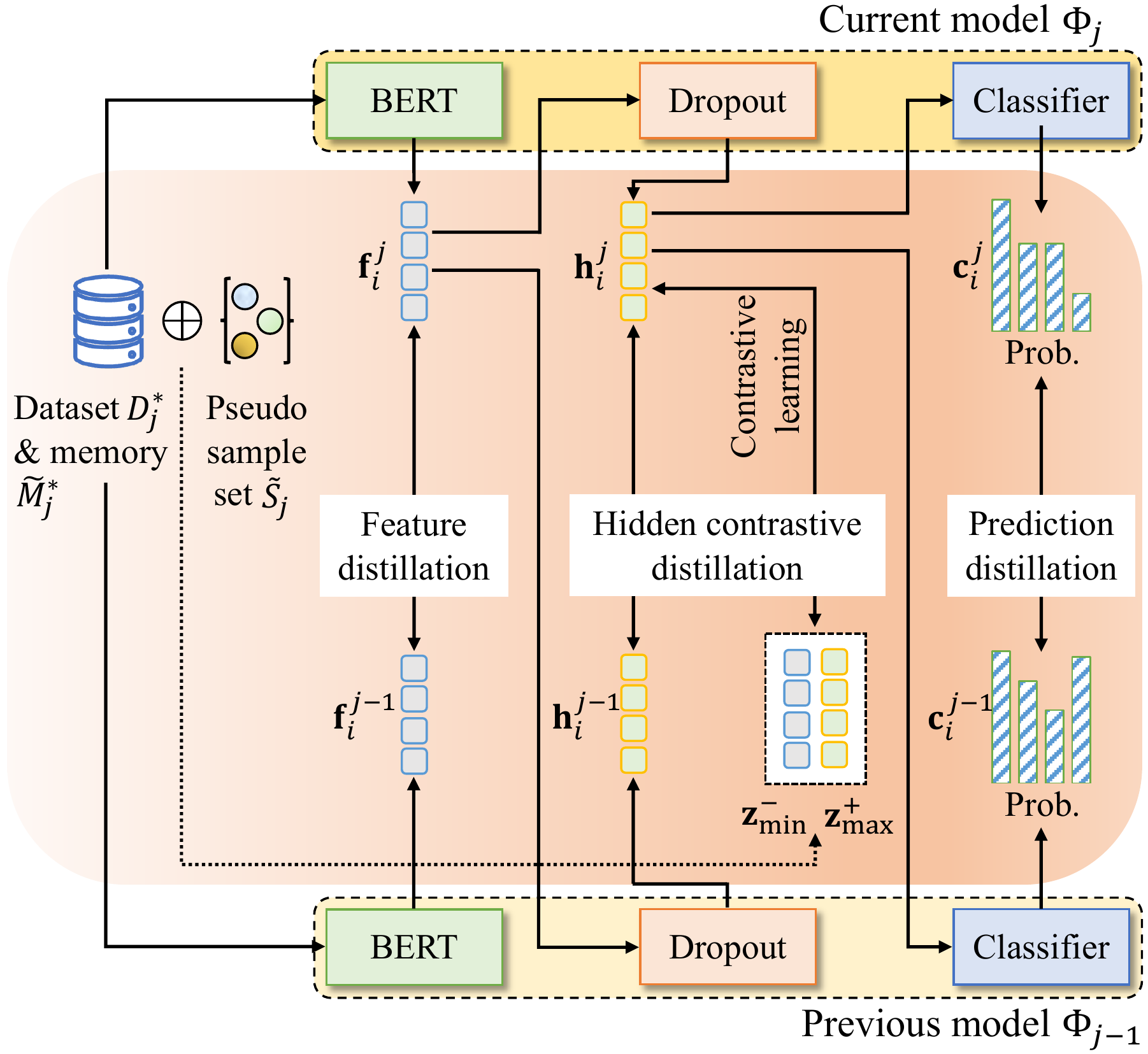}
\caption{Procedure of serial contrastive knowledge distillation.}
\label{fig:distillation}
\end{figure}

Knowledge distillation \cite{hinton2015distilling,cao2020incremental} has demonstrated its effectiveness in transferring knowledge. 
In this paper, we propose a serial contrastive knowledge distillation method to leverage the knowledge from the previous RE model to guide the training of the current model. 
The procedure of serial contrastive knowledge distillation is depicted in Figure~\ref{fig:distillation}.
We detail it below.

\paragraph{Feature distillation.}
In this step, we expect the encoder of the current model to extract similar features with the previous model.
For a sample $x$, let $\mathbf{f}_x^{j-1}$ and $\mathbf{f}_x^j$ be $x$'s features extracted by the previous model $\Phi_{j-1}$ and the current model $\Phi_j$, respectively. 
We propose a feature distillation loss to enforce the extracted features unbiased towards new relations:
\begin{align}
\mathcal{L}_\text{fd} = \frac{1}{|\tilde{M}_j^*|} \sum_{x \in \tilde{M}_j^*} \Big(1-({\mathbf{f}}_x^{j-1})^{\top}\mathbf{f}_x^j\Big).
\end{align}

\paragraph{Pseudo samples generation.}
We attempt to construct pseudo samples for all the observed relations, which are used in the next hidden contrastive distillation step.
Specifically, we assume the sample representations of relations follow the Gaussian distribution with the corresponding prototypes as their average values. 
The construction of pseudo samples is based on prototype set $\tilde{P}_j$, and one pseudo sample for $r$ can be constructed as follows:
\begin{align}
    \mathbf{s}_r = \mathbf{p}_r + \eta\cdot\delta_r,
\end{align}
where $\eta\sim\mathcal{N}(0,1)$ is a standard Gaussian noise, and $\delta_r$ is the root of the diagonal covariance based on the hidden representations of all $r$'s samples when $r$ first appears in the relation set of one task. 
The diagonal covariance consists of the variance in each dimension, which can describe the differences in each dimension of the sample representations belonging to that relation. 
We multiply the Gaussian noise with the root of the diagonal covariance and add the result to the prototype representation for generating pseudo samples. 
In this way, the generated samples can more closely match the real samples of the relation rather than random.
We repeat the above operation $n$ times for each relation in $\{T_1,\dots,T_j\}$ and store the constructed pseudo samples in the pseudo sample set $\tilde{S}_j$.

\paragraph{Hidden contrastive distillation.}
In this step, we expect the current model to obtain similar hidden representations with the previous model.
We also want to keep the hidden representations of samples in different relations distinguishable.

First, we consider the distillation between sample hidden representations. 
We feed a sample $x$'s feature $\mathbf{f}_x^j$ into the dropout layers of the previous model $\Phi_{j-1}$ and the current model $\Phi_j$ to obtain the hidden representations, which are denoted by $\mathbf{h}^{j-1}_x$ and $\mathbf{h}^j_x$, respectively. 
Then, we formulate the representation distillation loss as follows:
\begin{align}
\mathcal{L}_\text{rd} = \frac{1}{|\tilde{M}_j^*|} \sum_{x\in\tilde{M}_j^*} \Big(1-(\mathbf{h}_x^{j-1})^\top\mathbf{h}_x^j\Big).
\end{align}

Moreover, based on the previously-constructed pseudo samples and the real samples from the training data, we conduct contrastive learning to make the hidden representations of samples for different relations as distinct as possible, which can enhance the knowledge distillation. 
To achieve this, we mine hard positives and hard negatives with previous representations while contrasting them with current representations, which can ensure that the current model can obtain similar representations as the previous model. 
We put forward a distillation triplet loss function: 
\begin{align}
\begin{split}
\mathcal{L}_\text{dtr} = \frac{1}{|\tilde{M}_j^*|}\sum_{x\in\tilde{M}_j^*} & \max\Big(0, ||\mathbf{h}_x^j-\mathbf{z}^+_{\max}||_2\\
& - ||\mathbf{h}_x^j-\mathbf{z}^-_{\min}||_2\Big),
\end{split}
\end{align}
where $\mathbf{z}^+_{\max}$ and $\mathbf{z}^-_{\min}$ are selected through $\mathbf{h}^{j-1}_x$. 
$\mathbf{z}^+_{\max}$ is the representation farthest from $\mathbf{h}^{j-1}_x$ in all sample representations that belong to the same relation with $x$, and $\mathbf{z}^-_{\min}$ is the representation nearest from $\mathbf{h}^{j-1}_x$ in all sample representations that belong to the different relations with $x$.

Overall, the loss function for hidden contrastive distillation is defined as
\begin{align}
\mathcal{L}_\text{hcd} = \mathcal{L}_\text{rd} +\mathcal{L}_\text{dtr}.
\end{align}

\paragraph{Prediction distillation.}
In this step, we expect the classifier of the current model to predict similar probability distributions with the classifier of the previous model on the previous relation set.
For a sample $x$'s hidden representation $\mathbf{h}_x^j$, the output logits of the previous model are $\mathbf{o}^{j-1}_x = \Big[o^{j-1}_{x,1},\dots,o^{j-1}_{x,|\tilde{R}_{j-1}|}\Big]$ while the logits of the current model are $\mathbf{o}^j_x = \Big[o^j_{x,1},\dots,o^j_{x,|\tilde{R}_{j-1}|},\dots,o^j_{x,|\tilde{R}_j|}\Big]$. 
We propose a prediction distillation loss function:
\begin{align}
&\mathcal{L}_\text{pd} = -\frac{1}{|\tilde{M}_j^*|}\sum_{x\in\tilde{M}_j^*}\sum_{r=1}^{|\tilde{R}_{j-1}|} \mathbf{c}_{x,r}^{j-1} \log\mathbf{c}_{x,r}^j,\\
&\resizebox{\columnwidth}{!}{$
\mathbf{c}_{x,r}^{j-1} = \frac{\exp\Big(\frac{\mathbf{o}_{x,r}^{j-1}}{T}\Big)}{\sum_{l=1}^{|\tilde{R}_{j-1}|} \exp\Big(\frac{\mathbf{o}_{x,l}^{j-1}}{T}\Big)}, 
\mathbf{c}_{x,r}^j = \frac{\exp\Big(\frac{\mathbf{o}_{x,r}^j}{T}\Big)}{\sum_{l=1}^{|\tilde{R}_{j-1}|} \exp\Big(\frac{\mathbf{o}_{x,l}^j}{T}\Big)},
$}\end{align}
where $T$ is the temperature scalar. 
This prediction distillation loss encourages the predictions of the current model on previous relations to match the soft labels by the previous model.

The total distillation loss consists of the above three losses:
\begin{align}
\mathcal{L}_\text{dst} = \alpha\cdot\mathcal{L}_\text{fd} + \beta\cdot\mathcal{L}_\text{hcd} + \gamma\cdot\mathcal{L}_\text{pd},
\end{align}
where $\alpha,\beta$ and $\gamma$ are adjustment coefficients.

We optimize the classification loss and distillation loss with multi-task learning.
Therefore, the final loss is
\begin{align}
\mathcal{L} = \lambda_1 \cdot \mathcal{L}_\text{csf} + \lambda_2 \cdot \mathcal{L}_\text{dst},
\end{align}
where $\lambda_1$ and $\lambda_2$ are also adjustment coefficients.

%====================%
\section{Experiments}
\label{sect:experimens}

In this section, we assess the proposed SCKD and report our results. 
The datasets and source code for SCKD are accessible from GitHub.\footnote{\url{https://github.com/nju-websoft/SCKD}}

%--------------------%
\subsection{Experiment Setup}

\paragraph{Datasets.}
Our experiments are conducted on the following two benchmark RE datasets:
\begin{itemize}
\item\emph{FewRel} \cite{han2018fewrel} is a popular dataset for few-shot RE containing 100 relations and 70,000 samples in total. 
Following \cite{qin2022continual}, we adopt the version of 80 relations and split them into 8 tasks, where each task contains 10 relations (\textit{10-way}). 
The first task $T_1$ has 100 samples per relation while the subsequent tasks $T_2,\dots,T_8$ are all few-shot. 
We conduct \textit{5-shot} and \textit{10-shot} experiments.

\item\emph{TACRED} \cite{zhang2017position} is a large-scale RE dataset with 42 relations and 106,264 samples from Newswire and Web
documents. 
Following \cite{qin2022continual}, we filter out ``\textit{no\_relation}'' and divide the remaining 41 relations into 8 tasks. 
The first task $T_1$ has 6 relations and 100 samples per relation. 
All the other tasks have 5 relations (\textit{5-way}), and we carry out \textit{5-shot} and \textit{10-shot} experiments.
\end{itemize}

\paragraph{Evaluation metrics.}
We measure \emph{average accuracy} in our experiments.
At task $T_j$, it can be calculated as $\text{ACC}_j = \frac{1}{j}\sum_{i=1}^{j} \text{ACC}_{j,i}$,
where $\text{ACC}_{j,i}$ denotes the accuracy (i.e., the number of correctly-labeled samples divided by all samples) on the test set of task $T_i$ after training the model on task $T_j$. 
We repeat the experiments six times using random seeds, and report means and standard deviations.

\paragraph{Competing models.}

We compare SCKD against two baselines: 
The \emph{finetuning} model trains the RE model only with the training data of the current task while inheriting the parameters of the model trained on the previous task. 
It serves as the lower bound.
The \emph{joint-training} model stores all samples from previous tasks in memory and uses all the memorized data to train the re-initialized model for the current task. 
It can be regarded as the upper bound.

We also compare SCKD with four recent open-source models for continual RE: 
RP-CRE \cite{cui2021refining}, CRL \cite{zhao2022consistent}, CRECL \cite{hu2022improving}, and ERDA \cite{qin2022continual}.
Since RP-CRE, CRL, and CRECL do not investigate the few-shot scenario while ERDA reported its results under the ``loose'' evaluation which picks no more than 10 negative labels from the observed labels, we re-run these models using their source code and report the new results.
KIP-Framework \cite{zhang2022prompt} has not released its source code, thus we cannot re-run it for comparison.

\paragraph{Implementation details.}
We develop our SCKD based on PyTorch 1.7.1 and Huggingface's Transformers 2.11.0 \cite{wolf2020transformers}.
%The training procedure is optimized with Adam.
See Appendix~\ref{sec:EH} for the selected hyperparameter values.

For a fair comparison, we set the random seeds of the experiments identical to those in \cite{qin2022continual}, so that the task sequence is exactly the same. 
We employ the ``strict'' evaluation method proposed in \cite{cui2021refining}, which chooses the whole observed relation labels as negative labels for evaluation. 
We stipulate that the memory can only store one sample for each relation ($L=1$) when running all models. 

\begin{table*}
\centering
\resizebox{\textwidth}{!}{
\begin{tabular}{l|cccccccc}
\toprule
FewRel      & $T_1$ & $T_2$ & $T_3$ & $T_4$ & $T_5$ & $T_6$ & $T_7$ & $T_8$ \\
\midrule
Finetune    & $94.32_{\pm 0.21}$ & $43.54_{\pm 2.18}$ & $28.19_{\pm 1.51}$ & $22.46_{\pm 0.64}$ & $17.89_{\pm 0.92}$ & $14.39_{\pm 0.91}$ & $12.61_{\pm 0.65}$ & $10.68_{\pm 0.64}$ \\
Joint-train & $94.87_{\pm 0.27}$ & $80.83_{\pm 3.79}$ & $74.41_{\pm 2.32}$ & $71.73_{\pm 0.85}$ & $70.12_{\pm 2.55}$ & $67.37_{\pm 1.62}$ & $65.67_{\pm 1.75}$ & $64.48_{\pm 0.45}$ \\
\midrule
RP-CRE      & $93.97_{\pm 0.64}$ & $76.05_{\pm 2.36}$ & $71.36_{\pm 2.83}$ & $69.32_{\pm 3.98}$ & $64.95_{\pm 3.09}$ & $61.99_{\pm 2.09}$ & $60.59_{\pm 1.87}$ & $59.57_{\pm 1.13}$ \\
CRL         & $94.68_{\pm 0.33}$ & $80.73_{\pm 2.91}$ & $73.82_{\pm 2.77}$ & $70.26_{\pm 3.18}$ & $66.62_{\pm 2.74}$ & $63.28_{\pm 2.49}$ & $60.96_{\pm 2.63}$ & $59.27_{\pm 1.32}$ \\
CRECL       & $93.93_{\pm 0.22}$ & $82.55_{\pm 6.95}$ & $74.13_{\pm 3.59}$ & $69.33_{\pm  3.87}$ & $66.51_{\pm 4.05}$ & $64.60_{\pm 1.92}$ & $62.97_{\pm 1.46}$ & $59.99_{\pm 0.65}$ \\
ERDA        & $92.43_{\pm 0.32}$ & $64.52_{\pm 2.11}$ & $50.31_{\pm 3.32}$ & $44.92_{\pm 3.77}$ & $39.75_{\pm 3.34}$ & $36.36_{\pm 3.12}$ & $34.34_{\pm 1.83}$ & $31.96_{\pm 1.91}$ \\
\midrule
SCKD        & $94.77_{\pm 0.35}$ & $82.83_{\pm 2.61}$ & $76.21_{\pm 1.61}$ & $72.19_{\pm 1.33}$ & $70.61_{\pm 2.24}$ & $67.15_{\pm 1.96}$ & $64.86_{\pm 1.35}$ & $62.98_{\pm 0.88}$ \\
\bottomrule
\toprule
TACRED      & $T_1$ & $T_2$ & $T_3$ & $T_4$ & $T_5$ & $T_6$ & $T_7$ & $T_8$ \\
\midrule 
Finetune    & $87.97_{\pm 0.53}$ & $25.81_{\pm 4.57}$ & $19.65_{\pm 4.75}$ & $18.38_{\pm 1.25}$ & $15.68_{\pm 1.31}$ & $11.88_{\pm 2.61}$ & $10.77_{\pm 2.49}$ & $9.69_{\pm 2.26}$ \\
Joint-train & $87.93_{\pm 0.68}$ & $78.02_{\pm 1.51}$ & $72.84_{\pm 1.38}$ & $68.23_{\pm 5.21}$ & $63.42_{\pm 4.98}$ & $62.01_{\pm 3.89}$ & $59.62_{\pm 2.33}$ & $57.63_{\pm 1.41}$ \\
\midrule
RP-CRE      & $87.32_{\pm 1.76}$ & $74.90_{\pm 6.13}$ & $67.88_{\pm 4.31}$ & $60.02_{\pm 5.37}$ & $53.26_{\pm 4.67}$ & $50.72_{\pm 7.62}$ & $46.21_{\pm 5.29}$ & $44.48_{\pm 3.74}$ \\
CRL         & $88.32_{\pm 1.26}$ & $76.30_{\pm 7.48}$ & $69.76_{\pm 5.89}$ & $61.93_{\pm 2.55}$ & $54.68_{\pm 3.12}$ & $50.92_{\pm 4.45}$ & $47.00_{\pm 3.78}$ & $44.27_{\pm 2.51}$ \\
CRECL       & $87.09_{\pm 2.50}$ & $78.09_{\pm 5.74}$ & $61.93_{\pm 4.89}$ & $55.60_{\pm 5.78}$ & $53.42_{\pm 2.99}$ & $51.91_{\pm 2.95}$ & $47.55_{\pm 3.38}$ & $45.53_{\pm 1.96}$ \\
ERDA        & $81.88_{\pm 1.97 }$ & $53.68_{\pm 6.31}$ & $40.36_{\pm 3.35}$ & $36.17_{\pm 3.65}$ & $30.14_{\pm 3.96}$ & $22.61_{\pm 3.13}$ & $22.29_{\pm 1.32}$ & $19.42_{\pm 2.31}$ \\
\midrule
SCKD        & $88.42_{\pm 0.83}$ & $79.35_{\pm 4.13}$ & $70.61_{\pm 3.16}$ & $66.78_{\pm 4.29}$ & $60.47_{\pm 3.05}$ & $58.05_{\pm 3.84}$ & $54.41_{\pm 3.47}$ & $52.11_{\pm 3.15}$ \\
\bottomrule
\end{tabular}}
\caption{Result comparison on FewRel (10-way-5-shot) and TACRED (5-way-5-shot). $\text{Means}_{\,\pm\,\text{stds}}$ are reported.}
\label{tab:main}
\end{table*}

%--------------------%
\subsection{Results and Analyses}

\begin{table}
\centering
\resizebox{\columnwidth}{!}{
    \begin{tabular}{l|cccccccc}
    \toprule
    FewRel          & $T_1$ & $T_2$ & $T_3$ & $T_4$ & $T_5$ & $T_6$ & $T_7$ & $T_8$ \\
    \midrule
    SCKD            & 94.77 & 82.83 & 76.21 & 72.19 & 70.61 & 67.15 & 64.86 & 62.98 \\
    \ \ w/o dst.    & 94.67 & 82.47 & 74.13 & 68.59 & 66.31 & 63.43 & 61.36 & 58.96 \\
    \ \ w/o aug.    & 94.77 & 82.56 & 75.78 & 71.75 & 70.37 & 66.87 & 64.39 & 62.51 \\
    \ \ w/o both     & 94.63 & 82.39 & 73.96 & 68.14 & 65.97 & 62.92 & 60.62 & 58.41 \\
    \bottomrule
    \toprule
    TACRED          & $T_1$ & $T_2$ & $T_3$ & $T_4$ & $T_5$ & $T_6$ & $T_7$ & $T_8$ \\
    \midrule 
    SCKD            & 88.42 & 79.35 & 70.61 & 66.78 & 60.47 & 58.05 & 54.41 & 52.11 \\ 
    \ \ w/o dst.    & 88.38 & 77.12 & 66.95 & 61.64 & 56.25 & 53.39 & 48.09 & 46.52 \\
    \ \ w/o aug.    & 88.35 & 79.16 & 70.08 & 66.32 & 60.15 & 57.73 & 54.04 & 51.79 \\
    \ \ w/o both     & 88.12 & 76.48 & 65.45 & 60.99 & 55.79 & 52.46 & 47.31 & 45.79 \\
    \bottomrule
    \end{tabular}}
\caption{Ablation study on modules.}
\label{tab:ablation}
\end{table}

\subsubsection{Main Results}

Table \ref{tab:main} lists the result comparison on the 10-way-5-shot setting on the FewRel dataset and the 5-way-5-shot setting on the TACRED dataset.
We have the following findings:
(1) Our proposed SCKD performs significantly better than the competing models on all tasks. 
After learning all tasks, SCKD outperforms the second-best model CRECL by 2.99\% and 6.09\% on FewRel and TACRED, respectively.
(2) Regarding the two baselines, the finetuning model leads to rapid drops in average accuracy due to severe overfitting and catastrophic forgetting.
The joint-training model may not always be the upper bound (e.g., $T_2$ to $T_5$ on FewRel) due to the extremely imbalanced data distribution. 
Besides, after learning the final task of FewRel, SCKD can achieve close results to the joint-training model with considerably few memorized samples.
(3) ERDA performs worst among the four competing models. 
This is because the extra training data from the unlabeled Wikipedia corpus for data augmentation may contain errors and noise, which makes the model unable to fit the emerging relations well.
(4) RP-CRE, CRL, and CRECL can effectively acquire knowledge from new relations without catastrophic forgetting of prior knowledge. 
However, their performance is all affected by the limited memory size, since they all need more memorized samples for each relation to generate more representative relation prototypes.

See Appendix \ref{subsec:DS} for the 10-way-10-shot results on FewRel and 5-way-10-shot results on TACRED.

%--------------------%
\subsubsection{Ablation Study}

We conduct an ablation study to validate the effectiveness of each module. 
Specifically, for ``w/o distillation'', we disable the serial contrastive knowledge distillation module. 
For ``w/o augmentation'', we use the original (not augmented) dataset and memory. 
For ``w/o both'', we update the model via the simple re-training on memory.
From Table~\ref{tab:ablation}, we obtain several findings: 
(1) The average accuracy at each task reduces when we disable any modules, showing their usefulness.
(2) If we remove the serial contrastive knowledge distillation module, the results drop drastically, which shows that knowledge distillation and contrastive learning can alleviate catastrophic forgetting and overfitting.

\begin{table}
\centering
\resizebox{\columnwidth}{!}{
    \begin{tabular}{l|cccccccc}
    \toprule
    FewRel & $T_1$ & $T_2$ & $T_3$ & $T_4$ & $T_5$ & $T_6$ & $T_7$ & $T_8$ \\
    \midrule
    SCKD & 94.77 & 82.83 & 76.21 & 72.19 & 70.61 & 67.15 & 64.86 & 62.98 \\
    \ \ w/o $\mathcal{L}_\text{fd}$ & 94.75 & 82.78 & 75.95 & 71.89 & 69.93 & 66.74 & 64.04 & 62.59 \\
    \ \ w/o $\mathcal{L}_\text{rd}$ & 94.67 & 82.37 & 75.58 & 71.84 & 69.96 & 66.93 & 64.19 & 62.44 \\
    \ \ w/o $\mathcal{L}_\text{dtr}$ & 94.78 & 81.75 & 75.14 & 71.32 & 69.38 & 65.53 & 62.86 & 61.18 \\
    \ \ w/o $\mathcal{L}_\text{pd}$ & 94.71 & 82.12 & 75.48 & 71.75 & 69.91 & 66.47 & 63.95 & 61.98 \\
    \bottomrule
    \toprule
    TACRED & $T_1$ & $T_2$ & $T_3$ & $T_4$ & $T_5$ & $T_6$ & $T_7$ & $T_8$ \\
    \midrule 
    SCKD & 88.42 & 79.35 & 70.61 & 66.78 & 60.47 & 58.05 & 54.41 & 52.11 \\
    \ \ w/o $\mathcal{L}_\text{fd}$ & 88.38 & 79.07 & 70.25 & 66.51 & 59.95 & 57.89 & 53.18 & 51.59 \\
    \ \ w/o $\mathcal{L}_\text{rd}$ & 88.45 & 78.57 & 69.96 & 66.21 & 60.09 & 57.96 & 53.53 & 51.67 \\    \ \ w/o $\mathcal{L}_\text{dtr}$ & 88.37 & 78.36 & 69.78 & 65.21 & 59.81 & 57.46 & 52.85 & 50.94 \\
    \ \ w/o $\mathcal{L}_\text{pd}$ & 88.41 & 79.11 & 70.16 & 66.14 & 60.02 & 57.86 & 53.26 & 51.56\\
    \bottomrule
    \end{tabular}}
\caption{Fine-grained ablation study on serial contrastive knowledge distillation.}
\label{tab:distillation}
\end{table}

Furthermore, we conduct a fine-grained ablation study to investigate serial contrastive knowledge distillation. 
We disable $\mathcal{L}_\text{fd}, \mathcal{L}_\text{rd}, \mathcal{L}_\text{dtr}, \mathcal{L}_\text{pd}$ in the model update, to assess their influence. 
Table \ref{tab:distillation} shows the results, and we have several findings:
(1) The results decline if we remove any losses, which demonstrates that each loss contributes to the overall performance. 
(2) The drops caused by disabling the distillation triplet loss $\mathcal{L}_\text{dtr}$ are most obvious since SCKD cannot keep the hidden representations of samples in different relations sufficiently distinguishable without contrastive learning.

%--------------------%
\subsubsection{Comparison with Few-shot RE Models}

\begin{table}
\centering
\resizebox{\columnwidth}{!}{
    \begin{tabular}{l|ccccccc}
    \toprule
    FewRel & $T^*_2$ & $T^*_3$ & $T^*_4$ & $T^*_5$ & $T^*_6$ & $T^*_7$ & $T^*_8$ \\
    \midrule
    SCKD & 86.08 & 86.11 & 89.88 & 89.17 & 87.67 & 89.42 & 87.71 \\
    GNN (CNN) & \ \ 9.93 & \ \ 9.50 & \ \ 9.62 & \ \ 9.60 & \ \ 9.77 & \ \ 9.93 & 10.12 \\
    Proto (CNN) & 12.78 & 14.05 & 14.87 & 13.05 & 13.77 & 13.35 & 12.78 \\
    Proto (BERT) & 76.42 & 77.65 & 77.23 & 75.93 & 78.83 & 84.28 & 80.71 \\
    BERT-PAIR & 82.03 & 80.17 & 80.73 & 80.42 & 81.78 & 84.01 & 81.70 \\
    \bottomrule
    \toprule
    TACRED & $T^*_2$ & $T^*_3$ & $T^*_4$ & $T^*_5$ & $T^*_6$ & $T^*_7$ & $T^*_8$ \\
    \midrule 
    SCKD & 92.34 & 93.06 & 87.35 & 84.97 & 93.73 & 86.94 & 90.98 \\
    GNN (CNN) & 23.14 & 19.83 & 18.41 & 19.41 & 19.42 & 18.92 & 20.34 \\
    Proto (CNN) & 36.48 & 28.65 & 27.00 & 28.99 & 24.75 & 20.31 & 22.21 \\
    Proto (BERT) & 61.61 & 58.87 & 71.23 & 65.12 & 72.86 & 56.60 & 68.41 \\
    BERT-PAIR & 62.46 & 68.34 & 75.83 & 74.05 & 69.73 & 65.63 & 73.68 \\
    \bottomrule
    \end{tabular}}
\caption{Result comparison with few-shot RE models.}
\label{tab:fewshot}
\end{table}

We compare SCKD with classic few-shot RE models provided in \cite{gao2019fewrel}.
For a fair comparison, the few-shot RE models treat the training and test sets of previous tasks as the support and query sets for training, respectively. 
The training set of the current task serves as the support set for testing.
We test our model and the few-shot models using the accuracy on the test set of \emph{current} task.

Table \ref{tab:fewshot} presents the results, and we observe that SCKD is always superior to GNN (CNN), Proto (CNN), Proto (BERT), and BERT-PAIR, as it conducts contrastive learning with pseudo samples on the few-shot tasks, which maximizes the distance between the representations of different relations.

%\begin{figure}
%\centering
%\includegraphics[width=\columnwidth]{fewshot.pdf}
%\caption{Results comparison with few-shot RE models.}
%\label{fig:fewshot}
%\end{figure}

%--------------------%
\subsubsection{Knowledge Transfer Capability}

\emph{Backward transfer} (BWT) measures how well the continual learning model can handle catastrophic forgetting. 
The BWT of accuracy after finishing all tasks is defined as follows:
\begin{align}
\text{BWT} = \frac{1}{J-1}\sum_{i=1}^{J-1}\Big(\text{ACC}_{J,i}-\text{ACC}_{i,i}\Big).
\end{align}

\begin{figure}
\centering
\includegraphics[width=\columnwidth]{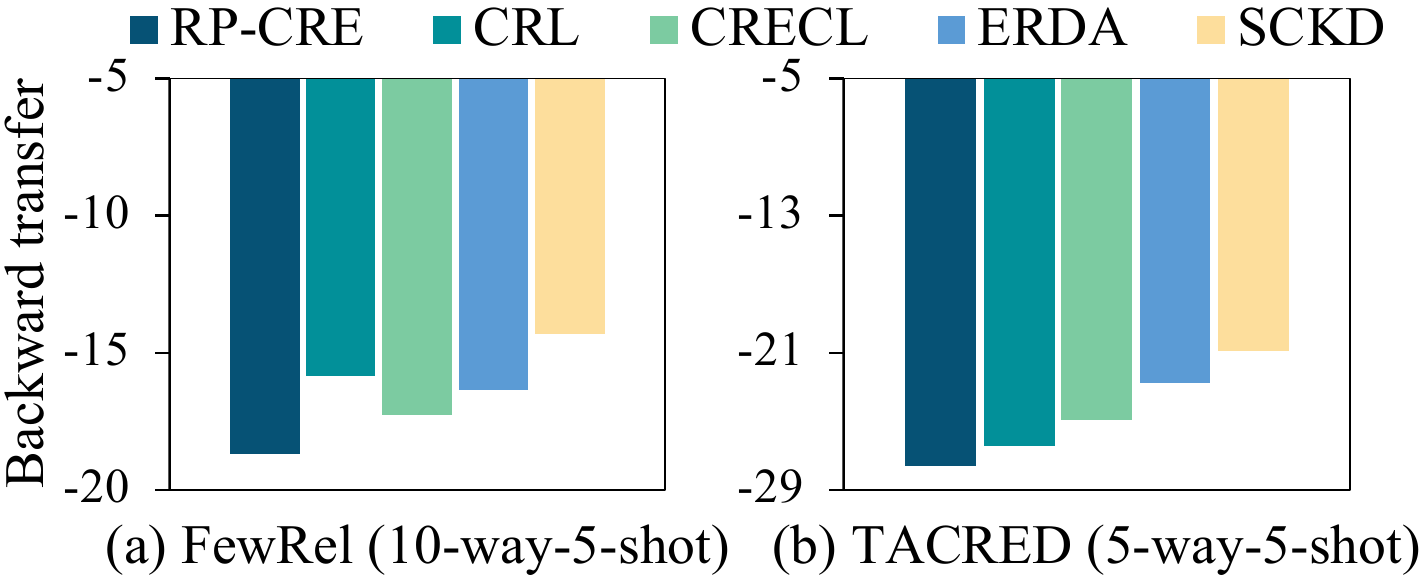}
\caption{Results of BWT on FewRel and TACRED.}
\label{fig:BWT}
\end{figure}

Figure \ref{fig:BWT} shows the BWT of SCKD and the competing models. 
Due to the overwriting of learned knowledge, BWT is always negative. 
The performance drops of SCKD are the lowest, showing its effectiveness in alleviating catastrophic forgetting.
See Appendix \ref{subsec:KTC} for the 10-shot results on FewRel and TACRED.

%--------------------%
\subsubsection{Sample Representation Discrimination}
To investigate the effects on discriminating sample representations, we use t-SNE \cite{van2008visualizing} to visualize the sample representations of six selected relations after the training of CRECL and SCKD.

\begin{figure}
\centering
\includegraphics[width=.95\columnwidth]{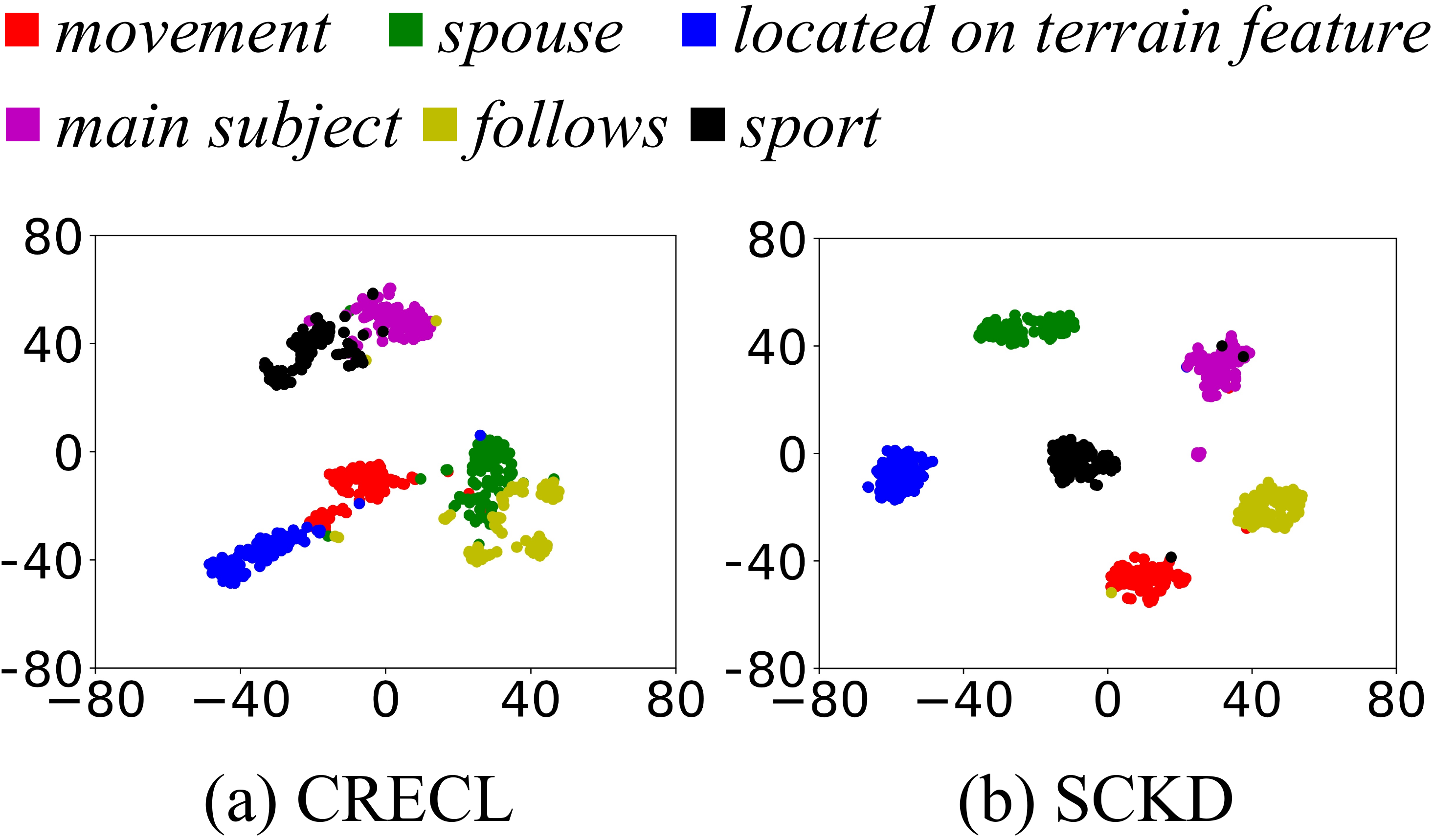}
\caption{t-SNE plot of sample representations belonging to six relations from FewRel (10-way-5-shot)}
\label{fig:tsne}
\end{figure}

From Figure \ref{fig:tsne}, we see that, compared to CRECL, SCKD can make the representations of samples in different relations more distinguishable. 
For example, the two relations, ``\textit{spouse}'' and ``\textit{follows}'', with close sample representations in CRECL can be clearly separated by SCKD, which shows that SCKD has a better ability to maintain the differences between relations.

%--------------------%
\subsubsection{Influence of Memory Size}

For the memory-based continual RE models, memory size has an important impact on performance. 
Due to the limited samples in the few-shot scenario, the models only store one sample per relation ($L=1$) in the previous experiments.
In this experiment, we conduct experiments on the 10-way-10-shot setting of FewRel with different memory sizes ($L=2,3$). 
We choose this setting because it ensures that the memorized data only occupy a small fraction of all samples.

The comparison results are shown in Table \ref{tab:memory}, and we can see that: 
(1) With memory size growing, all the models perform better, confirming that memory size is a key factor that affects continual learning.
(2) SCKD maintains the best performance with different memory sizes, which demonstrates the effectiveness of SCKD in leveraging the memory for continual few-shot RE.
See Appendix \ref{subsec:IMS} for the results on TACRED.

\begin{table}
\centering
\resizebox{\columnwidth}{!}{
    \begin{tabular}{l|cccccccc}
    \toprule
    $L=2$ & $T_1$ & $T_2$ & $T_3$ & $T_4$ & $T_5$ & $T_6$ & $T_7$ & $T_8$ \\
    \midrule
    RP-CRE      & 95.22 & 83.27 & 79.62 & 75.84 & 73.86 & 70.12 & 69.06 & 67.41 \\
    CRL         & 95.21 & 84.21 & 80.97 & 76.77 & 74.49 & 71.44 & 69.39 & 67.03 \\
    CRECL       & 95.21 & 85.82 & 80.09 & 76.27 & 74.13 & 71.91 & 70.21 & 67.89 \\
    ERDA        & 92.67 & 68.63 & 61.64 & 55.69 & 47.81 & 43.72 & 41.91 & 39.80 \\
    SCKD        & 95.25 & 87.83 & 81.56 & 77.59 & 75.91 & 73.04 & 70.96 & 68.82 \\
    \bottomrule
    \toprule
    $L=3$ & $T_1$ & $T_2$ & $T_3$ & $T_4$ & $T_5$ & $T_6$ & $T_7$ & $T_8$ \\
    \midrule
    RP-CRE      & 94.88 & 84.73 & 82.67 & 79.79 & 74.73 & 71.96 & 71.05 & 69.31 \\
    CRL         & 95.11 & 85.32 & 81.46 & 79.65 & 76.14 & 73.28 & 72.12 & 69.85 \\
    CRECL       & 95.16 & 86.26 & 81.25 & 77.63 & 75.52 & 74.01 & 72.05 & 69.99 \\
    ERDA        & 92.81 & 70.60 & 63.18 & 59.09 & 51.46 & 47.72 & 45.33 & 43.51 \\
    SCKD        & 95.27 & 88.63 & 83.21 & 80.13 & 77.18 & 75.15 & 73.22 & 71.04 \\
    \bottomrule
    \end{tabular}}
\caption{Results w.r.t. memory size on FewRel (10-way-10-shot).}
\label{tab:memory}
\end{table}

%====================%
\section{Conclusion}
\label{sect:conclusion}

In this paper, we propose SCKD for continual few-shot RE. 
To alleviate the problems of catastrophic forgetting and overfitting, we design the serial contrastive knowledge distillation, making prior knowledge from previous models sufficiently preserved while the representations of samples in different relations remain distinguishable. 
Our experiments on FewRel and TACRED validate the effectiveness of SCKD for continual few-shot RE and its superiority in knowledge transfer and memory utilization. 
For future work, we plan to investigate how to apply the serial contrastive knowledge distillation to other classification-based continual few-shot learning tasks.

%====================%
\section{Limitations}

The work presented here has a few limitations: 
(1) The proposed model belongs to the memory-based methods for continual learning, which requires a memory that costs extra storage. 
In some extremely storage-sensitive cases, there may be restrictions on the usage of our model.
(2) The proposed model has currently been evaluated under the RE setting.
It is better to transfer it to other continual few-shot learning settings (e.g., event detection and even image classification) for a comprehensive study.

%====================%
\section*{Acknowledgments}

This work was supported by the National Natural Science Foundation of China (No. 62272219) and the Collaborative Innovation Center of Novel Software Technology \& Industrialization.

%====================%
\balance
\bibliography{custom}
\bibliographystyle{acl_natbib}

\clearpage
\appendix

% \section{Example Appendix}
% \label{sec:appendix}

% This is a section in the appendix.

%====================%
\section{Environment and Hyperparameters}
\label{sec:EH}

We run all the experiments on an X86 server with two Intel Xeon Gold 6326 CPUs, 512 GB memory, four NVIDIA RTX A6000 GPU cards, and Ubuntu 20.04 LTS. 
The training procedure is optimized with Adam.
Following the convention, we conduct a grid search to choose the hyperparameter values.
Specifically, the search space of important hyperparameters is as follows:
\begin{enumerate}
\item The search range for the dropout ratio is $[0.2,0.6]$ with a step size of 0.1.
\item The search range for the threshold $\tau$ is $[0.80,0.99]$ with a step size of 0.01.
\item The search range for the number of pseudo samples per relation is $[5,20]$ with a step size of 5.
\item The search range for $\alpha, \beta, \gamma$ and $\lambda_1, \lambda_2$ is $[0.1,1]$ with a step size of 0.1.
\end{enumerate}

The selection is illustrated in Table~\ref{tab:hyper}.
\begin{table}[!h]
\centering
\resizebox{\columnwidth}{!}{
    \begin{tabular}{l|c}
    \toprule
    Hyperparameters & Values \\ 
    \midrule
    Batch size & $16$ \\
    Dropout ratio & $0.5$ \\
    Gradient accumulation steps & $4$ \\
    Learning rate for the encoder & $0.00001$ \\
    Learning rate for the dropout layer & $0.00001$ \\
    Learning rate for the classifier & $0.001$ \\
    \midrule
    Dim. of BERT representations & $768$ \\
    Dim. of hidden representations & $768$ \\
    Threshold $\tau$ for augmentation & $0.95$ \\
    No. of pseudo samples per relation & $10$ \\
    Temperature scalar & $0.08$ \\
    $\alpha,\beta,\gamma$ & $0.5, 1.0, 0.5$ \\
    $\lambda_1,\lambda_2$ & $1, 1$ \\
    \bottomrule
    \end{tabular}}
\caption{Hyperparameter setting in our model.}
\label{tab:hyper}
\end{table}

For all the competing models ERDA \cite{qin2022continual}, RP-CRE \cite{cui2021refining}, CRL \cite{zhao2022consistent} and CRECL \cite{hu2022improving}, we just assign the same memory size as ours, and retain other hyperparameter settings reported in their original papers.

%====================%
\section{More Experimental Results}
\label{sec:MER}

%--------------------%
\subsection{Knowledge Transfer Capability}
\label{subsec:KTC}

\begin{figure}[!h]
\centering
\includegraphics[width=\columnwidth]{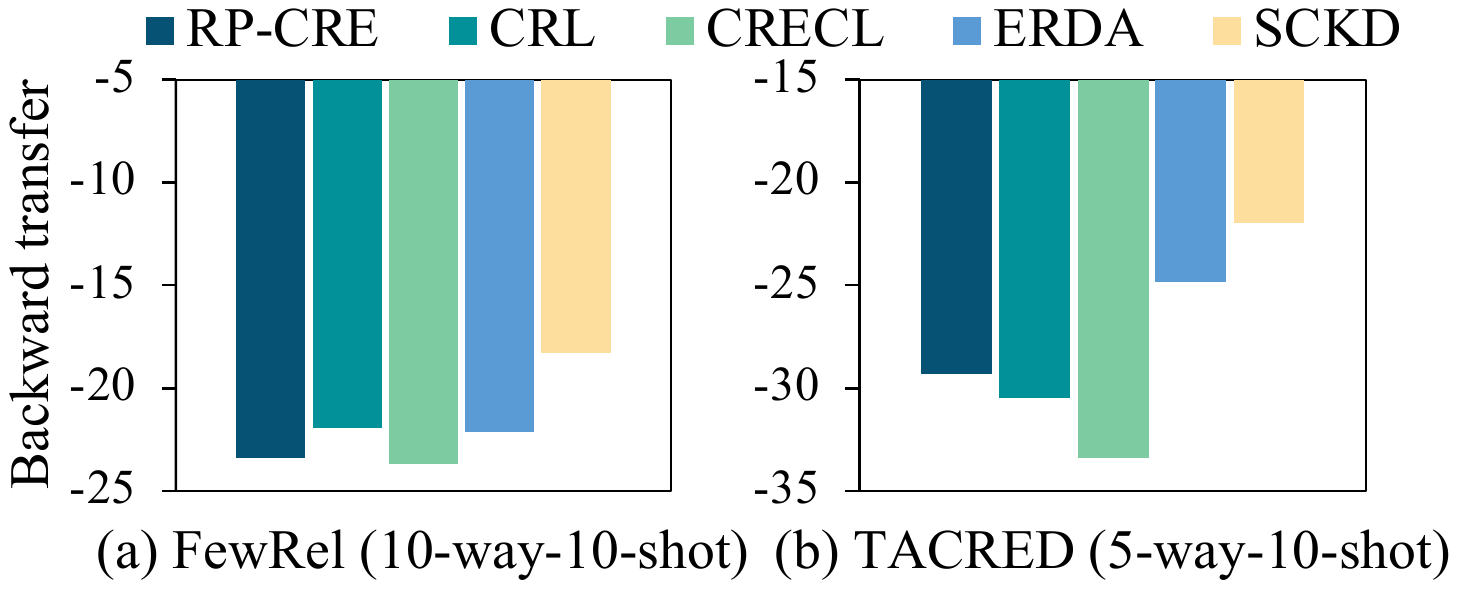}
\caption{Results of BWT on FewRel (10-way-10-shot) and TACRED (5-way-10-shot).}
\label{fig:BWT10shot}
\end{figure}

Figure \ref{fig:BWT10shot} presents the 10-way-10-shot BWT results on FewRel and the 5-way-10-shot BWT results on TACRED. 
From this figure as well as Figure \ref{fig:BWT} in the main text, we can observe that: 
(1) SCKD achieves the best BWT scores again under this different shot setting. 
(2) Compare with the competing models, the performance of SCKD declines lowest, which shows that SCKD alleviates catastrophic forgetting effectively.

%--------------------%
\subsection{Influence of Memory Size}
\label{subsec:IMS}

To enrich the experimental results on the influence of memory size, we also conduct an experiment on TACRED with different memory sizes and show the results in Table \ref{tab:tacredshot}. 
Based on these results and the results listed in Table~\ref{tab:memory} of the main text, we can find that: 
SCKD maintains the best performance with different memory sizes not only on FewRel but also on TACRED. 
This demonstrates that our model is effective and versatile in making good use of memory.

\begin{table}[!h]
\centering
\resizebox{\columnwidth}{!}{
    \begin{tabular}{l|cccccccc}
    \toprule
    $L=2$ & $T_1$ & $T_2$ & $T_3$ & $T_4$ & $T_5$ & $T_6$ & $T_7$ & $T_8$ \\
    \midrule
    RP-CRE  &   86.42 & 78.69 & 70.41 & 62.73 & 58.37 & 55.79 & 52.55 & 50.43 \\
    CRL     &   86.71 & 77.48 & 68.02 & 61.65 & 59.18 & 56.55 & 53.45 & 52.18 \\
    CRECL   &   84.58 & 74.83 & 66.80 & 57.57 & 56.58 & 55.26 & 52.26 & 52.01 \\
    ERDA    &   79.69 & 54.06 & 40.40 & 34.41 & 33.34 & 29.47 & 28.43 & 26.51 \\
    SCKD    &   88.27 & 79.07 & 71.11 & 64.88 & 62.14 & 58.91 & 56.41 & 54.84 \\
    \bottomrule
    \toprule
    $L=3$ & $T_1$ & $T_2$ & $T_3$ & $T_4$ & $T_5$ & $T_6$ & $T_7$ & $T_8$ \\
    \midrule
    RP-CRE  &   87.19 & 78.98 & 70.57 & 63.25 & 60.68 & 57.24 & 55.78 & 51.89 \\
    CRL     &   87.01 & 79.35 & 69.94 & 62.96 & 61.01 & 58.72 & 56.61 & 53.76 \\
    CRECL   &   86.06 & 76.93 & 68.39 & 62.83 & 60.11 & 59.78 & 56.81 & 55.96 \\
    ERDA    &   80.75 & 55.13 & 44.63 & 37.29 & 34.53 & 32.37 & 31.13 & 29.20 \\
    SCKD    &   88.59 & 80.47 & 74.26 & 66.56 & 64.85 & 61.78 & 59.34 & 56.74 \\
    \bottomrule
    \end{tabular}}
\caption{Results w.r.t. memory size on TACRED (5-way-10-shot).}
\label{tab:tacredshot}
\end{table}

\begin{table*}
\centering
\resizebox{\textwidth}{!}{
    \begin{tabular}{l|cccccccc}
    \toprule
    FewRel & $T_1$ & $T_2$ & $T_3$ & $T_4$ & $T_5$ & $T_6$ & $T_7$ & $T_8$ \\
    \midrule
    Finetune    & $95.67_{\pm 0.28}$ & $46.64_{\pm 2.22}$ & $29.68_{\pm 1.98}$ & $22.41_{\pm 1.48}$ & $18.47_{\pm 0.58}$ & $14.84_{\pm 0.99}$ & $13.02_{\pm 0.59}$ & $11.23_{\pm 0.72}$ \\
    Joint-train & $95.82_{\pm 0.37}$ & $87.17_{\pm 5.11}$ & $80.73_{\pm 5.95}$ & $77.75_{\pm 5.33}$ & $76.77_{\pm 3.74}$ & $74.26_{\pm 2.14}$ & $72.96_{\pm 1.81}$ & $71.57_{\pm 0.39}$ \\
    \midrule
    RP-CRE      & $95.19_{\pm 0.21}$ & $79.21_{\pm 6.35}$ & $74.72_{\pm 4.18}$ & $71.39_{\pm 5.11}$ & $67.62_{\pm 3.83}$ & $64.43_{\pm 2.72}$ & $63.08_{\pm 2.59}$ & $61.46_{\pm 1.19}$ \\
    CRL         & $95.01_{\pm 0.31}$ & $82.08_{\pm 6.91}$ & $79.52_{\pm 4.85}$ & $75.48_{\pm 4.91}$ & $69.41_{\pm 3.05}$ & $66.49_{\pm 2.23}$ & $64.86_{\pm 1.45}$ & $62.95_{\pm 0.59}$ \\
    CRECL       & $95.63_{\pm 0.28}$ & $83.81_{\pm 3.69}$ & $78.06_{\pm 5.91}$ & $71.28_{\pm 4.54}$ & $68.32_{\pm 3.52}$ & $66.76_{\pm 3.84}$ & $64.95_{\pm 1.40}$ & $63.01_{\pm 1.62}$ \\
    ERDA        & $92.68_{\pm 0.57}$ & $66.59_{\pm 8.29}$ & $56.33_{\pm 6.23}$ & $48.62_{\pm 5.96}$ & $40.51_{\pm 2.22}$ & $37.21_{\pm 2.25}$ & $36.39_{\pm 3.17}$ & $33.51_{\pm 1.47}$ \\
    \midrule
    SCKD        & $95.45_{\pm 0.34}$ & $86.64_{\pm 4.72}$ & $80.06_{\pm 6.73}$ & $76.02_{\pm 5.96}$ & $73.82_{\pm 4.33}$ & $70.57_{\pm 3.22}$ & $68.34_{\pm 2.34}$ & $66.66_{\pm 0.75}$ \\
    \bottomrule
    \toprule
    TACRED & $T_1$ & $T_2$ & $T_3$ & $T_4$ & $T_5$ & $T_6$ & $T_7$ & $T_8$ \\
    \midrule 
    Finetune    & $85.84_{\pm 1.95}$ & $25.63_{\pm 3.75}$ & $21.49_{\pm 4.63}$ & $17.45_{\pm 2.05}$ & $14.32_{\pm 1.95}$ & $13.14_{\pm 3.01}$ & $11.34_{\pm 2.59}$ & $9.21_{\pm 1.59}$ \\
    Joint-train & $86.56_{\pm 1.12}$ & $80.14_{\pm 2.17}$ & $74.67_{\pm 2.86}$ & $70.31_{\pm 2.79}$ & $70.04_{\pm 2.96}$ & $67.31_{\pm 2.19}$ & $65.42_{\pm 2.03}$ & $61.59_{\pm 1.19}$ \\
    \midrule
    RP-CRE      & $86.68_{\pm 1.72}$ & $78.43_{\pm 4.25}$ & $69.43_{\pm 6.22}$ & $60.71_{\pm 4.34}$ & $55.84_{\pm 5.28}$ & $51.17_{\pm 4.24}$ & $47.27_{\pm 3.49}$ & $47.16_{\pm 1.88}$ \\
    CRL         & $87.81_{\pm 0.39}$ & $77.68_{\pm 7.89}$ & $63.31_{\pm 7.77}$ & $56.51_{\pm 2.82}$ & $53.21_{\pm 2.01}$ & $52.42_{\pm 4.02}$ & $48.54_{\pm 4.19}$ & $46.46_{\pm 3.73}$ \\
    CRECL       & $83.88_{\pm 1.68}$ & $73.45_{\pm 2.85}$ & $59.24_{\pm 5.55}$ & $53.51_{\pm 5.04}$ & $49.27_{\pm 3.24}$ & $47.41_{\pm 2.85}$ & $45.15_{\pm 3.61}$ & $44.33_{\pm 2.48}$ \\
    ERDA        & $79.37_{\pm 0.95}$ & $51.28_{\pm 5.67}$ & $36.97_{\pm 4.95}$ & $29.39_{\pm 5.07}$ & $27.80_{\pm 4.23}$ & $25.18_{\pm 3.29}$ & $24.47_{\pm 1.22}$ & $22.37_{\pm 3.92}$ \\
    \midrule
    SCKD        & $88.84_{\pm 1.51}$ & $78.64_{\pm 5.03}$ & $70.08_{\pm 3.17}$ & $64.27_{\pm 2.99}$ & $61.73_{\pm 2.82}$ & $58.19_{\pm 3.95}$ & $55.91_{\pm 2.79}$ & $52.95_{\pm 3.14}$ \\
    \bottomrule
    \end{tabular}}
\caption{Result comparison on FewRel (10-way-10-shot) and TACRED (5-way-10-shot). $\text{Means}_{\,\pm\,\text{stds}}$ are reported.}
\label{tab:othershot}
\end{table*}

\subsection{Results with Different Shots}
\label{subsec:DS}

Table \ref{tab:othershot} shows the 10-way-10-shot results on the FewRel dataset and the 5-way-10-shot results on the TACRED dataset.
Based on these results and the experimental results on memory size listed in Table~\ref{tab:memory} of the main text, we have the following findings: 
(1) Compared with the competing models, our model still performs best.
It gains a significant accuracy improvement over the second-best model by 3.65\% on FewRel and 5.79\% on TACRED at last. 
(2) Our model achieves a close performance with $L = 1$ (62.98\% on FewRel and 52.11\% on TACRED) to the competing models with $L = 2$. This demonstrates that our model can make better use of memory.

\end{document}